\documentclass[preprint,12pt]{elsarticle}

\usepackage{longtable}
\usepackage[utf8]{inputenc} 
\usepackage[T1]{fontenc}    
\usepackage{hyperref}       
\usepackage{url}            
\usepackage{booktabs}       
\usepackage{amsfonts}       
\usepackage{nicefrac}       
\usepackage{microtype}      
\usepackage{amsmath}
\usepackage{xcolor}
\usepackage{lipsum}
\usepackage{makecell}
\usepackage{adjustbox}
\usepackage{multirow}
\usepackage{tabularx}

\journal{Medical Image Analysis}

\begin{document}

\begin{frontmatter}

\title{Comparison of different segmentation algorithms on brain volume and fractal dimension in infant brain MRIs}
\author[label1,label2]{Nathalie Alexander}
\author[label3,label4]{Arnaud Gucciardi}
\author[label3,label2]{Umberto Michelucci}

\affiliation[label1]{organization={Laboratory for Motion Analysis, Devision of Paediatric Orthopaedic, Children's Hospital of Eastern Switzerland},
            city={St. Gallen},
            country={Switzerland}}
\affiliation[label2]{organization={Lucerne University of Applied Sciences and Arts},
            city={Lucerne},
            country={Switzerland}}
\affiliation[label3]{organization={TOELT llc, Machine Learning Research and Development},
            city={Dübendorf},
            country={Switzerland}}
\affiliation[label4]{organization={University of Ljubljana, Faculty of Computer and Information Science},
            city={Ljubljana},
            country={Slovenia}}

\begin{abstract}
Accurate segmentation of infant brain MRI is essential for quantifying developmental changes in structure and complexity. However, ongoing myelination and reduced tissue contrast make automated segmentation particularly challenging. This study systematically compared segmentation accuracy and its impact on volumetric and fractal dimension (FD) estimates in infant brain MRI using the Baby Open Brains (BOB) dataset (71 scans, 1–9 months). Two methods, SynthSeg and SamSeg, were evaluated against expert annotations using Dice, Intersection over Union, 95th-percentile Hausdorff distance, and Normalised Mutual Information. SynthSeg outperformed SamSeg across all quality metrics (mean Dice > 0.8 for major regions) and provided volumetric estimates closely matching the manual reference (mean +4\% [-28\% - 71\%]). SamSeg systematically overestimated ventricular and whole-brain volumes (mean +76\% [-12\% - 190\%]). Segmentation accuracy improved with age, consistent with increasing tissue contrast during myelination. FD analyses revealed significant regional differences between SynthSeg and expert segmentations, and Bland–Altman limits of agreement indicated that segmentation-related FD variability exceeded most group differences reported in developmental cohorts. Volume and FD deviations were positively correlated across structures, indicating that segmentation bias directly affects FD estimation. Overall, SynthSeg provided the most reliable volumetric and FD results for paediatric MRI, yet small morphological differences in volume and FD should be interpreted with caution due to segmentation-related uncertainty.
\end{abstract}

\begin{graphicalabstract}
\end{graphicalabstract}

\begin{highlights}
\item SynthSeg yields more accurate infant brain MRI segmentations than SamSeg

\item Segmentation reliability increases with age and ongoing myelination

\item Segmentation bias jointly distorts regional brain volume and fractal dimension

\item Segmentation uncertainty can exceed reported developmental fractal dimension effects
\end{highlights}

\begin{keyword}
Fractal Dimension \sep Neonatal \sep Segmentation \sep Pediatric
\end{keyword}

\end{frontmatter}


\section{Introduction}

The human brain is a complex organ, with several regions~\cite{nieuwenhuys2007human}. Magnetic resonance imaging (MRI) is widely used to investigate brain structures and tissue composition in vivo. The T1-weighted sequences provide high anatomical detail, with cerebrospinal fluid (CSF) appearing dark and fat bright, making them suitable for structural evaluation. In contrast, T2-weighted imaging highlights pathology, as the CSF appears bright and lesions such as edoema or demyelination are more noticeable due to longer T2 relaxation times~\cite{westbrook2018mri}. Within these modalities, brain segmentation is a critical step that forms the basis for subsequent analyses such as cortical surface reconstruction and volumetric quantification. However, processing infant brain magnetic resonance images is particularly challenging compared to adults, as ongoing myelination leads to reduced tissue contrast, age-dependent intensity changes, and regionally heterogeneous appearances, compounded by small brain size and partial volume effects. Myelination begins before birth and continues to increase throughout the first three years of life~\cite{carmody2004quantitative,parazzini2002terminal}. A recent review~\cite{defaria2021periods} suggests that myelination is an ongoing process throughout life and that myelin changes can be divided into five different stages during human life: early childhood, childhood, adolescence, adult myelination, and age-related decline in myelin.

Brain development in childhood is characterised by dynamic and region-specific volumetric changes. The total volume of the brain reaches about 95\% of adult size at age six~\cite{giedd2015child, lenroot2007sexual}, with grey matter following an inverted U-shaped trajectory and white matter continuing to increase until early adulthood due to myelination~\cite{lenroot2007sexual}. Individual structures show distinct patterns, such as non-linear hippocampal growth~\cite{uematsu2012developmental} or age-related ventricular enlargement, which is also diagnostically relevant in hydrocephalus~\cite{xenos2002ventricular,cutler2020normal}. Altered volumes have been reported across neurodevelopmental conditions, including enlarged brains in autism~\cite{courchesne2001unusual}, reduced cerebral and cerebellar volumes in attention-deficit/hyperactivity disorder~\cite{castellanos2002developmental}, and smaller brainstem structures in cerebral palsy~\cite{TREVARROW2023114}. Beyond brain volumes, fractal dimension (FD) has emerged as a sensitive marker of structural complexity. Introduced by Mandelbrot~\cite{mandelbrot1975stochastic,mandelbrot1983fractal}, FD captures irregularity and has been applied to study cognition, neurological disorders~\cite{DiMarco2024,DiazBeltran2024,Meregalli2022}, tumour progression~\cite{LegariaPena2023}, and age-related changes~\cite{Farahibozorg2015,Reishofer2018,Kalmanti2007}. Although most of the work has focused on adults, paediatric studies report age-related FD trajectories, hemispheric asymmetries, and sensitivity to early neurodevelopmental alterations~\cite{Kalmanti2007,Blanton2001,esteban2010fractal,sandu2014post}. Together, these findings highlight the need to examine both volumetric and fractal measures across different brain regions. Therefore, accurate segmentation is essential, motivating the comparison of different segmentation methods in paediatric brain MRI.

Accurate quantification of brain structure is essential for understanding both typical and atypical trajectories. Volumetric measures and emerging metrics such as FD offer valuable insight into neurodevelopmental patterns~\cite{Kalmanti2007,Blanton2001}. However, their reliability is highly dependent on the quality of the image~\cite{carr1991practice} and the subsequent segmentation. Brain segmentation can be performed manually by expert raters or automatically using deep learning-based software tools or methods. Manual segmentation is time-consuming and depends on anatomical expertise, but is considered the gold standard~\cite{darrault2025road}, however, it is not suitable for larger neuroimaging studies. On the other hand, automatic segmentation tools, such as FreeSurfer, FMRIB Software Library (FSL) or Statistical Parametric Mapping (SPM), and machine learning algorithms based on convolutional neural network (CNN) architectures have been developed over the years~\cite{singh2021review}. Various segmentation methods have been developed to address these challenges~\cite{puonti2016fast, billot2023synthseg, cardoso2022monai}, using spatial transformation and more recently deep learning models, each method offering unique strengths. 
Further, it has to be considered that FD estimation is also influenced by preprocessing choices, including image resolution, registration, binarization, and skeletonization~\cite{krohn2019evaluation}, while automated grid-scale optimisation methods have been shown to improve estimation accuracy~\cite{marzi2020toward}. Given these dependencies, consistent acquisition and processing pipelines are essential for obtaining reproducible results.

To the best of our knowledge, this study presents the first systematic evaluation of the impact of automated segmentation methods on both volumetric and fractal dimension (FD) estimates specifically within the challenging context of infant brain MRI. The main contributions of this work include: (i) a comprehensive benchmark of modern ``out-of-the-box'' segmentation algorithms (SynthSeg and SamSeg) against expert annotations using the Baby Open Brains (BOB) dataset, evaluating performance across varying stages of early myelination without the need for model retraining; (ii) a quantification of the dependency of segmentation accuracy on chronological age, offering insights into how rapid developmental changes in tissue contrast affect the reliability of automated processing in the first year of life; (iii) a demonstration of the propagation of segmentation errors into higher-order morphological biomarkers, establishing a positive correlation between volumetric bias and deviations in FD estimation; and (iv) the establishment of quantitative benchmarks for segmentation uncertainty through Bland--Altman limits of agreement, providing a critical reference framework for interpreting whether observed clinical or developmental differences in pediatric cohorts exceed inherent methodological variability.

\section{Methods}

\subsection{Dataset}

In this study, the Open Baby Brains (BOB) dataset~\cite{feczko2024baby} was used. In the dataset anatomical magnetic resonance imaging (MRI) data was obtained and segmented from 71 infant imaging visits across 51 participants (the mean age at the time of imaging was 5.3 months with a range of 1.0–9.4 months), with both T1w and T2w images. Detailed age distribution can be found in Table \ref{tab:demographics}.
 The choice of this dataset was due to the dramatic differences in myelination and intensities between ages in the dataset. This highlights the need of good segmentation in such a highly variable dataset and thus is a perfect testbed for our comparisons.

The BOB dataset provides high-quality MRI scans along with expert-validated segmentations, which serve as gold standards for evaluating other segmentation methods.

\begin{table}[ht]
\centering
\caption{Demographic characteristics of the children in the dataset, grouped by age category. For each group, the total number of images, sex distribution (female/male), mean chronological age (in months $\pm$ SD), and mean gestational age at birth (in weeks $\pm$ SD) are reported.}
\label{tab:demographics}
\begin{adjustbox}{max width=\textwidth}
\begin{tabular}{lcccc}
\toprule
\textbf{Age Category} &
\makecell[c]{\textbf{Total}\\\textbf{Count}} &
\makecell[c]{\textbf{Sex}\\(f / m)} &
\makecell[c]{\textbf{Chronological Age}\\\textbf{(months $\pm$ SD)}} &
\makecell[c]{\textbf{Gestational Age}\\\textbf{(weeks $\pm$ SD)}} \\
\midrule
1--2 months & 6 & 3 / 3 & 1.7 $\pm$ 0.3 & 40.7 $\pm$ 0.9 \\
2--3 months & 9 & 4 / 5 & 2.5 $\pm$ 0.3 & 40.3 $\pm$ 0.7 \\
3--4 months & 7 & 4 / 3 & 3.4 $\pm$ 0.2 & 39.9 $\pm$ 1.3 \\
4--5 months & 8 & 5 / 3 & 4.5 $\pm$ 0.3 & 40.2 $\pm$ 1.1 \\
5--6 months & 12 & 9 / 3 & 5.5 $\pm$ 0.2 & 39.8 $\pm$ 1.2 \\
6--7 months & 9 & 5 / 4 & 6.5 $\pm$ 0.2 & 38.6 $\pm$ 0.9 \\
7--8 months & 6 & 4 / 2 & 7.3 $\pm$ 0.1 & 39.6 $\pm$ 1.3 \\
8--9 months & 12 & 9 / 3 & 8.5 $\pm$ 0.3 & 39.8 $\pm$ 0.9 \\
9--10 months & 2 & 0 / 2 & 9.2 $\pm$ 0.3 & 40.2 $\pm$ 0.1 \\
\bottomrule
\end{tabular}
\end{adjustbox}
\end{table}

\subsection{Segmentation methods} 

To evaluate the impact of segmentation methods on volumetric and FD estimation in paediatric brain magnetic resonance imaging, we compared three state-of-the-art approaches to segmentation provided with the BOB data set. The methods tested are \textit{SynthSeg}, \textit{SamSeg}, and MONAI \textit{UNest}. They were applied to T1 and T2-weighted images of the 51 infants (71 scans). An overview of the methods tested can be found in Table \ref{tab:methods_summary}. An additional method, FastSurfer \cite{henschel2020fastsurfer}, was tested, but produced results that were not usable, and therefore was excluded from the analysis in this work.

\begin{table}[ht]
\centering
\caption{Summary of segmentation methods applied to the Baby Open Brains (BOB) dataset, including references, input modalities, and specific application details.}
\label{tab:methods_summary}
\begin{adjustbox}{max width=\textwidth}
\begin{tabular}{l p{4cm} c p{7cm}}
\toprule
\textbf{Segmentation Name} & \textbf{Reference} & \textbf{Modality} & \textbf{Application Details} \\
\midrule
SynthSeg & Billot et al. \cite{billot2023synthseg, billot2023robust} & T1w & Applied to T1w images without pre-processing (e.g., skull-stripping or normalization). CSF label was subtracted from the whole-brain segmentation to exclude non-brain regions. \\ \addlinespace
SamSeg & Puonti et al. \cite{puonti2016fast} & T1w + T2w & Applied using both T1w and T2w images (joint segmentation). CSF excluded for consistency with SynthSeg. Utilizes a Bayesian generative model. \\ \addlinespace
FastSurfer & Henschel et al. \cite{henschel2020fastsurfer} & T1w & Applied to the dataset but produced aberrant artifacts and anatomically implausible outputs; excluded from further analysis. \\ \addlinespace
MONAI UNest & Cardoso et al. \cite{cardoso2022monai}; Yu et al. \cite{yu2022characterizing} & T1w & Pre-trained model applied using standard hyperparameters without retraining. Resulted in highly unreliable segmentations and was excluded from analysis. \\
\bottomrule
\end{tabular}
\end{adjustbox}
\end{table}

\subsubsection{SynthSeg}
SynthSeg is a deep learning-based tool designed for robust brain segmentation across diverse MRI contrasts and resolutions. Using synthetic data generation, the model is exposed to a wide range of simulated contrasts, resolutions, noise levels, and artefacts. This domain randomisation process produces robust brain segmentation in various MRI contrasts and resolutions~\cite{billot2023synthseg,billot2023robust}.
This strong training approach ensures segmentation stability in various anatomies and MRI modalities, ideal for variability conditions in neonatal MRI. In addition to its robustness, SynthSeg comes with a fast processing speed and does not require pre-processing steps like skull-stripping or intensity normalization~\cite{billot2023robust}. This simplifies the workflow and reduces potential source of errors. The original study~\cite{billot2023synthseg} achieved high Dice scores in numerous public datasets ranging from 0.76 to 0.88.

In our implementation, SynthSeg was applied to the T1-weighted images provided in the dataset. To enable whole-brain comparisons, the cerebrospinal fluid (CSF) label was subtracted from the SynthSeg whole-brain segmentation to exclude non-brain regions. SynthSeg includes the CSF within its global brain mask, whereas the BOB annotations do not include extracerebral CSF. Therefore, CSF voxels were removed using a voxel-wise subtraction before computing whole-brain volume, ensuring consistent anatomical definitions across methods.

\subsubsection{SamSeg}
SAMSEG (Sequence Adaptive Multimodal SEGmentation) is a generative model–based segmentation tool developed within the FreeSurfer framework~\cite{puonti2016fast}. Unlike purely discriminative deep learning methods, SamSeg relies on a Bayesian generative modelling approach, in which both the observed MRI intensities and the segmentation labels are modelled probabilistically. This allows the method to account for uncertainty, adapt to different MRI contrasts, and simultaneously perform image inhomogeneity correction during segmentation. 

One of the key advantages of SamSeg is its ability to perform joint segmentation across multiple MRI sequences (e.g., T1-weighted, T2-weighted, FLAIR) without requiring retraining or manual parameter tuning. This makes it well suited for heterogeneous datasets and for studies involving pathological populations, where tissue characteristics and imaging protocols may vary substantially. The algorithm estimates probabilistic tissue priors in conjunction with subject-specific bias field correction, thus improving robustness across scanners and acquisition protocols.

In our study, SamSeg was applied using T1 and T2-weighted images from the BOB dataset. The segmentation outputs included cortical and subcortical structures, as well as tissue classes including gray matter, white matter, and the CSF. For whole-brain comparisons, we excluded the CSF part to maintain consistency with the SynthSeg outputs. Although SamSeg requires longer processing time compared to SynthSeg, its probabilistic framework provides valuable uncertainty estimates alongside the segmentations, which can be useful for quality control and downstream comparative analyses.

\subsubsection{FastSurfer}
Using the widely used FreeSurfer~\cite{fischl2012freesurfer,fischl2002whole} suite, FastSurfer~\cite{henschel2020fastsurfer} is designed to perform comprehensive cortical and subcortical segmentation and has been extensively validated in studies of adult population. FastSurfer enhances the base segmentation process by leveraging deep learning to accelerate segmentation while maintaining accuracy. 
We applied FastSurfer to the BOB dataset; however, the method produced aberrant artefacts and anatomically implausible outputs. This is likely due to its reliance on intensity distributions and optimised anatomical priors for adult brains, which do not account for the unique characteristics of infant MRI (e.g., ongoing myelination and reduced tissue contrast). Given these poor results, FastSurfer was excluded from further analysis.

\subsubsection{MONAI UNest}
Additionally, we evaluated an openly available 3D U-Net model provided on the MONAI platform~\cite{cardoso2022monai}. 
MONAI offers a suite of deep learning tools, including implementations of U-Net architectures, widely adopted for medical image segmentation tasks due to their encoder–decoder design and the ability to capture both global context and fine localisation. 

For our experiments, we applied a pre-trained UNEST model \cite{yu2022characterizing} for volumetric whole-brain segmentation on the T1-weighted images of the BOB dataset, using the standard hyperparameters and inference settings provided in the MONAI documentation. The UNEST model employs a “shifted window” transformer architecture to extend U-Net designs, allowing multi-scale feature modelling and efficient attention mechanisms. In principle, this architecture is well suited for handling complex structures in medical imaging. 

However, in practice, the resulting segmentations on our dataset proved to be highly unreliable. We observed substantial inaccuracies in both cortical and subcortical structures, with outputs often failing to capture brain boundaries or producing anatomically implausible results and completely different label maps. These inaccuracies are likely due to the fact that we did not retrain the model for our specific dataset, as this was beyond the scope of the present work.
Given the poor quality and lack of reproducibility of the segmentations, we did not include the UNEST outputs in subsequent analyses. Although promising in design, our findings suggest that this standard model is not currently suitable for robust neonatal brain segmentation without retraining or domain-specific adaptation.

\begin{figure}[ht]
    \centering
    \includegraphics[width=0.95\textwidth]{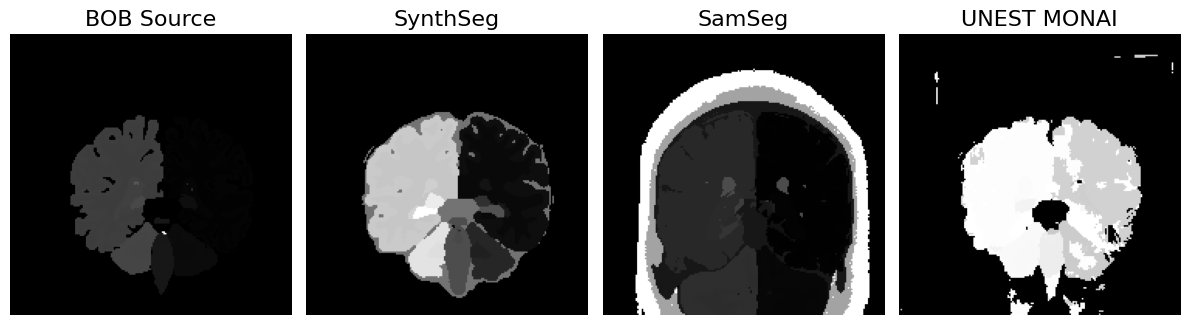}
    \caption{Qualitative comparison of segmentation results on a representative infant brain scan. The figure displays coronal slices of the segmentation label maps: (Left to Right) the expert-annotated ground truth (BOB Source), SynthSeg, SamSeg, and MONAI UNEST. Note the variations in tissue classification and boundary definition across the different algorithms.}
    \label{fig:segmentation}
\end{figure}

\subsection{Data analysis}

In order to quantify how differences in segmentation propagate to volumetric and fractal-dimension estimates, we performed a series of analyses at both the whole-brain and regional level. The primary goal of this pipeline was twofold: first, to characterise the agreement between automated methods and the expert-annotated reference for each anatomical structure; and second, to determine how segmentation-related variability influences the reliability of derived biomarkers such as volume and FD. All analyses were carried out on the same set of infant scans, ensuring that any observed discrepancies could be attributed to the segmentation approach rather than differences in image acquisition or sample composition.

We structured the data analysis into three stages. In the first stage, we evaluated the quality of the segmentation using established overlap and distance metrics, providing a benchmark comparison between SynthSeg and SamSeg for each labelled structure. In the second stage, we derived  meaningful quantities, regional volumes, and fractal dimensions, from the resulting label maps, and examined how these measures differed between methods and across hemispheres. Finally, we used a series of statistical tests and agreement analyses to relate segmentation quality to infant age, to quantify the coupling between volume and FD differences, and to establish limits of agreement that can serve as practical thresholds when interpreting developmental or clinical effects in future studies.

\subsubsection{Metrics}
To evaluate the performance of different brain segmentation methods, we used standard quantitative metrics including the Dice similarity coefficient~\cite{dice1945measures}, Intersection over Union (IoU)~\cite{jaccard1912distribution}, 95th percentile Hausdorff distance (HD95)~\cite{baid2021rsna, bouget2022preoperative}, and Normalised Mutual Information (NMI)~\cite{yokoi2004accuracy}. These metrics were chosen for their complementary strengths in assessing volumetric overlap, boundary alignment, and label agreement. 

The Dice similarity coefficient ~\cite{dice1945measures} quantifies the spatial overlap between a predicted segmentation region $A$ and a ground truth segmentation region $B$. It is defined as:

\begin{equation}
\text{Dice}(A, B) = \frac{2|A \cap B|}{|A| + |B|}
\end{equation}

In this study, $A$ denotes the automated segmentated region (either SynthSeg or SamSeg), and $B$ represents the expert-annotated reference region from the BOB dataset. $|A|$ and $|B|$ represent the total number of voxels in $A$ and $B$, respectively. $|A \cap B|$ is the number of voxels common to both $A$ and $B$.
The Dice score is widely used in brain magnetic resonance segmentation due to its intuitive interpretation~\cite{taha2015metrics}.

The IoU~\cite{jaccard1912distribution}, also known as the Jaccard index, measures the ratio of the intersection to the union of two segmentations:

\begin{equation}
\text{IoU}(A, B) = \frac{|A \cap B|}{|A \cup B|}
\end{equation}

Where $|A \cup B|$ represents the total number of voxels in either $A$ or $B$.
Both Dice and IoU assess volumetric agreement and range from 0 (no overlap) to 1 (perfect overlap), but IoU penalises discrepancies more strongly. It is particularly useful for evaluating segmentation accuracy in regions with complex boundaries or small structures.

The Hausdorff distance~\cite{baid2021rsna, bouget2022preoperative} evaluates the boundary deviation between two segmentations. To reduce sensitivity to outliers, we used the 95th percentile Hausdorff distance (HD95), defined as:

\begin{equation}
\text{HD}_{95}(A, B) = \max \left\{ \operatorname{percentile}_{95}\left( \min_{b \in B} \|a - b\| \right),\; \operatorname{percentile}_{95}\left( \min_{a \in A} \|b - a\| \right) \right\}
\end{equation}

With
\begin{itemize}
    \item $\partial A$ and $\partial B$ : sets of voxels on the boundaries of $A$ and $B$
    \item $\| a -b\|$ : the Euclidian distance between a voxel $a$ on the boundary of $A$ and a voxel $b$ on the boundary of $B$.
    \item $\operatorname{percentile}_{95}$ : The 95th percentile of distances.
\end{itemize}

The HD95 metric, expressed in millimetres, ranges from 0 (for a perfect boundary match) to arbitrarily large values, since it has no fixed upper bound. Its value depends on the resolution of the image and the extent of the boundary misalignment.

Lastly, NMI measures label agreement between two label distributions, in multi-class segmentations. It is defined as :

\begin{equation}
\text{NMI}(A, B) = \frac{2 I(A; B)}{H(A) + H(B)}
\end{equation}

where $I(A; B)$ is the mutual information, quantifying the shared information between two segmentations. $H(\cdot)$ is entropy, representing the uncertainty of each segmentation. 
 Normalisation restricts the NMI value to the range [0,1], where 0 indicates that there is no shared information and 1 represents perfect agreement. The normalised form is scale-independent and enables fair comparison across structures of different sizes or anatomical variability~\cite{studholme1999overlap,pluim2003mutual}.

\subsection{Volume and fractal dimension}
For the whole brain and for each segmented structure, volume and FD were calculated. CSF labels were excluded from the whole-brain segmentations to restrict the analyses to parenchymal tissue. 
 Volumes were derived directly from the segmentation masks by multiplying the number of voxels belonging to each label by the corresponding voxel volume, derived from the image header’s voxel dimensions (in mm$^3$). The resulting volumes were then converted from cubic millimetres to millilitres (mL).
 
 FD was calculated on three-dimensional binary masks of each segmented structure using the box-counting algorithm implemented in FractalBrain~\cite{marzi2020toward}.

\subsection{Statistical analysis}
In a first step, the above mentioned quality metrics were compared between SynthSeg and SamSeg for the whole brain and for each structure. 
Furthermore differences between left and right hemispheres were assessed to determine whether data could be pooled across hemispheres. Comparisons using paired $t$-test were performed within subjects for both volume and FD measures using the expert-annotated dataset as reference. Effect sizes were quantified with Cohen’s $d$ (which expresses the standardised difference between group means in standard deviation units), with thresholds defined as small ($d$ = 0.20–0.49), medium ($d$ = 0.50–0.79), or large ($d$ > 0.80)~\cite{cohen1992power}. To contextualise these hemispheric differences relative to segmentation variability, we compared left–right discrepancies with segmentation-related differences between expert annotations and automated segmentations.

Next, whole-brain and regional volumes were compared between expert annotations, SynthSeg, and SamSeg using repeated-measures ANOVA ($\alpha$ = 0.05). In case of significance, post-hoc tests were performed using pairwise paired $t$-tests with Bonferroni correction. Effect sizes were calculated as Cohen’s $d$~\cite{cohen1992power}.

For the best-performing segmentation method, additional analyses were performed. First, Pearson correlations ($r$) between age and quality metrics were calculated for each segmented structure. Furthermore, FD of the entire brain and segmented structures was compared between expert annotations and SynthSeg using paired $t$-tests, with effect sizes reported as Cohen’s $d$.

To examine the relationship between deviations in FD and volume, we computed correlations between FD differences and volume differences across segmented structures. FD and volume difference was defined as the signed difference between expert annotations and SynthSeg. For each structure, robust regression lines were fitted using the Theil–Sen estimator. To quantify associations, Spearman’s rank correlation coefficients ($\rho$) were calculated, as it is less sensitive to outliers. Outliers were identified for descriptive purposes using a median absolute deviation (MAD)–based criterion, in which a $z$-score was calculated by scaling the distance from the median by the median absolute deviation. Observations with an absolute $z$-score greater than 3.5 were considered extreme~\cite{iglewicz1993volume}. These points were reported, but not excluded, from the analysis.
\begin{table}[h!]
\centering
\footnotesize 
\caption{Summary of statistical analyses performed to evaluate segmentation methods.}
\label{tab:stat_analysis}
\renewcommand{\arraystretch}{1.3} 
\begin{tabular}{p{0.3\linewidth} p{0.3\linewidth} p{0.3\linewidth}}
\toprule
\textbf{Objective} & \textbf{Statistical Test / Method} & \textbf{Metrics \& Effect Sizes} \\ 
\midrule

\textbf{Hemispheric Comparison} \newline (Left vs. Right validation) & 
Paired $t$-test (within subjects). & 
Cohen's $d$ (Small: 0.2, Med: 0.5, Large: 0.8). \\ 
\addlinespace

\textbf{Method Comparison} \newline (Expert vs. SynthSeg vs. SamSeg) & 
Repeated-measures ANOVA ($\alpha = 0.05$). \newline \textit{Post-hoc:} Pairwise paired $t$-tests (Bonferroni correction). & 
Cohen's $d$ (Small: 0.2, Med: 0.5, Large: 0.8).  \\ 
\addlinespace

\textbf{Age Effects} \newline (Age vs. Quality Metrics) & 
Pearson correlation ($r$). & 
Correlation coefficient ($r$). \\ 
\addlinespace

\textbf{Fractal Dimension (FD)} \newline (Expert vs. SynthSeg) & 
Paired $t$-test. & 
Cohen's $d$ (Small: 0.2, Med: 0.5, Large: 0.8).  \\ 
\addlinespace

\textbf{Error Relationship} \newline (FD diff. vs. Volume diff.) & 
Spearman’s rank correlation ($\rho$). \newline Robust regression (Theil–Sen estimator). \newline Outlier detection via MAD ($z > 3.5$). & 
Spearman's $\rho$. \\ 
\addlinespace

\textbf{Agreement Analysis} \newline (Expert vs. SynthSeg) & 
Bland–Altman Analysis. & 
Mean Bias, 95\% Limits of Agreement (LoA). \\ 
\bottomrule
\end{tabular}
\end{table}

The agreement between SynthSeg and the expert-annotated reference was assessed using Bland–Altman analysis for each structure and for the whole brain. For each structure, the mean bias (defined as the average difference between SynthSeg and BOB) was calculated together with the 95\% limits of agreement (LoA), which were determined as the mean bias $\pm$ 1.96 times the standard deviation (SD) of the differences. An overview of the statistical analysis done can be found in Table \ref{tab:stat_analysis}.

\section{Results}

\begin{figure}
    \centering
    \includegraphics[width=0.8\linewidth]{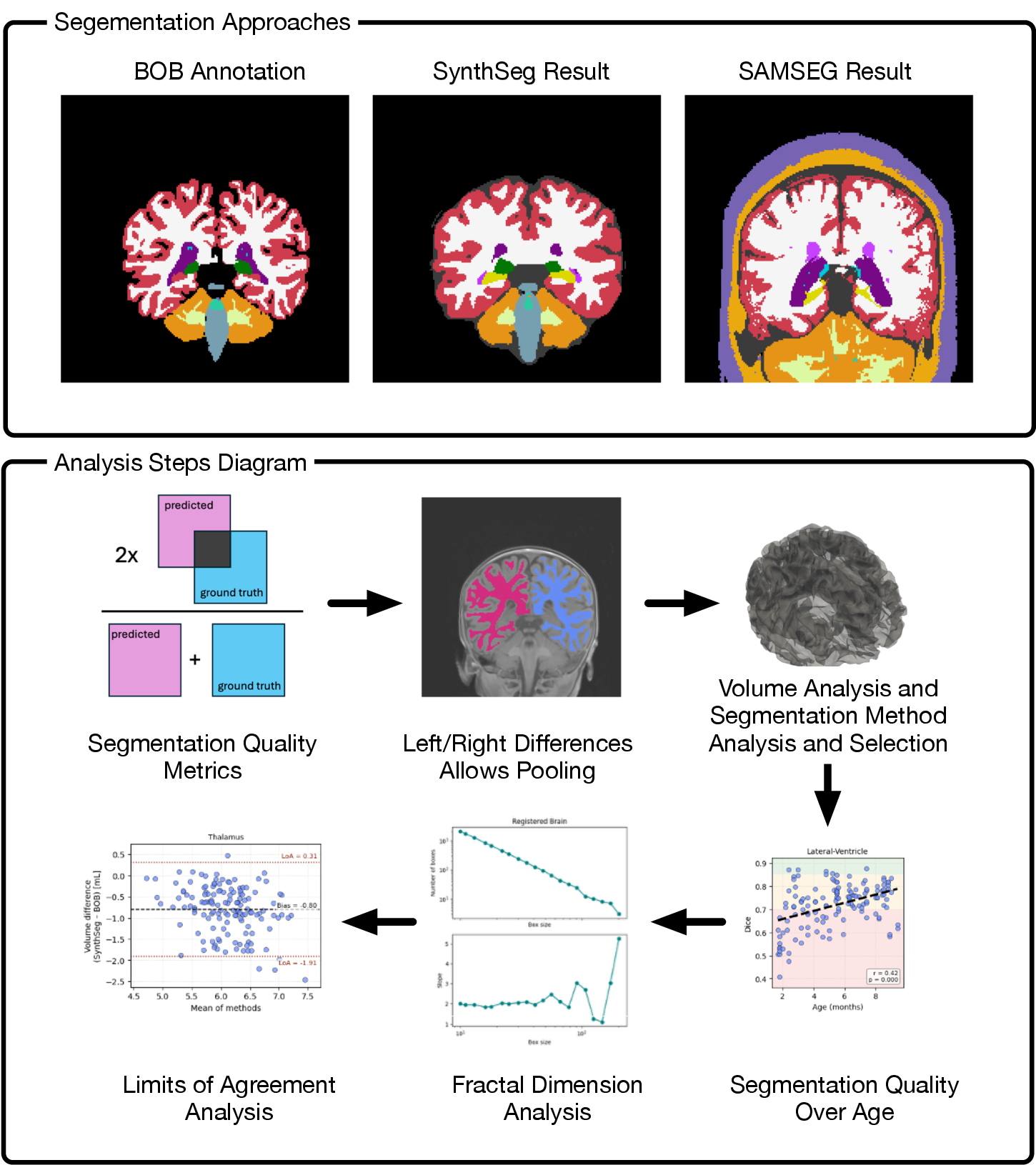}
    \caption{Overview of the segmentation approaches and analysis pipeline. \textbf{Top:} Example coronal slices illustrating the  segmentation label and approaches used in this study: expert Baby Open Brains (BOB) annotations (labels), SynthSeg, and SamSeg. \textbf{Bottom:} Schematic of the subsequent analysis workflow. Automated segmentations are first compared against the BOB reference using  quality metrics (Dice, IoU, HD95, NMI). Left–right differences in volume and fractal dimension (FD) are then evaluated to justify pooling hemispheres. Based on quality metrics and volume differences the better-performing method is selected for further analyses. Using this method, we examine age-related changes in segmentation quality, quantify segmentation-induced differences in FD (including the relationship with volume), and assess agreement of volume and FD with the expert reference via Bland–Altman analysis.}

    \label{fig:workflow}
\end{figure}

The analytical workflow is summarised in Figure \ref{fig:workflow}. The analyses were carried out sequentially, starting with an assessment of the quality of the segmentation, the evaluation of the hemispheric symmetry, and the comparison of the volumetric and fractal dimension (FD) metrics between the methods, followed by agreement analyses.

\subsection{Segmentation quality between SynthSeg and SamSeg}

\begin{figure}
    \centering
    \small
    \includegraphics[width=1\linewidth]{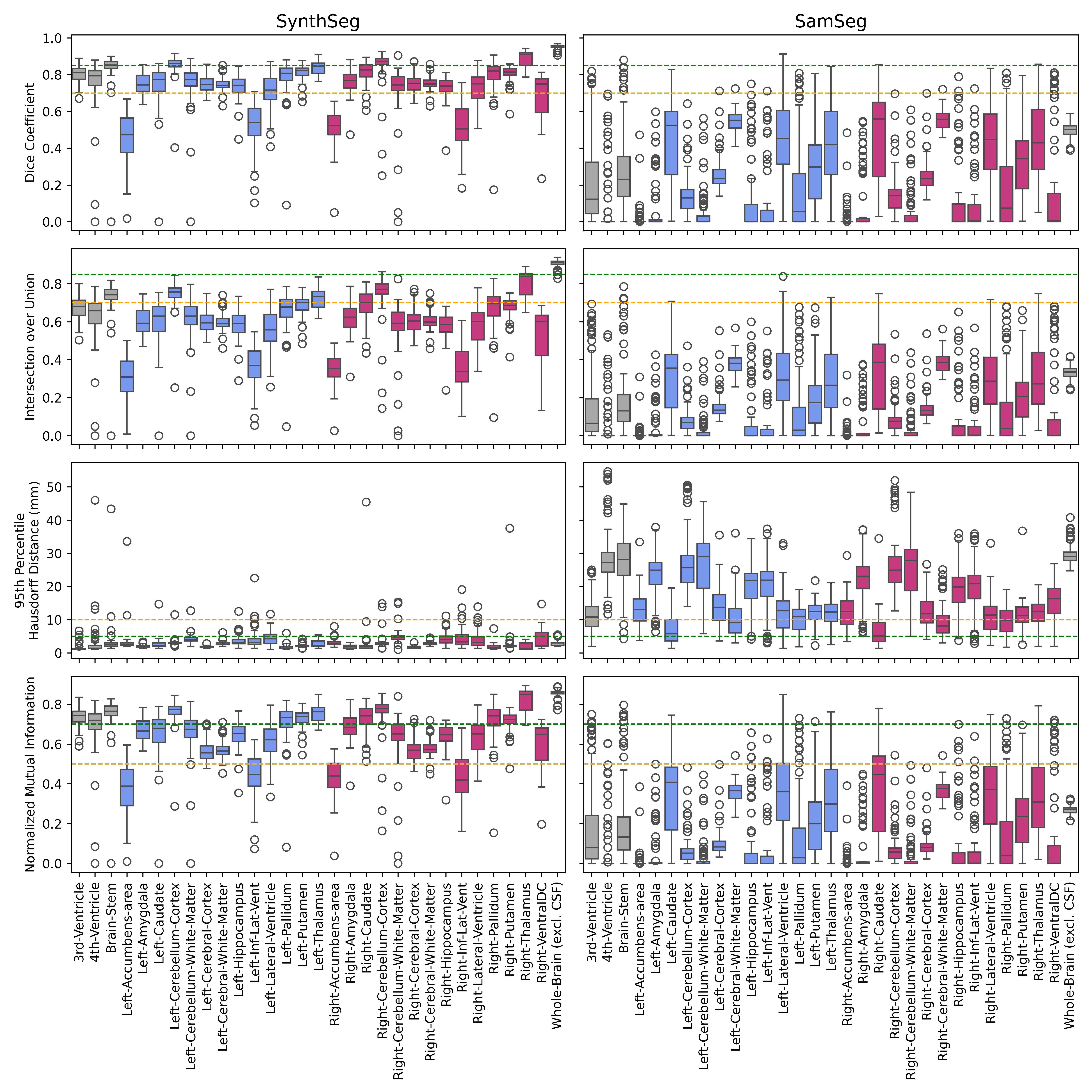}
    \caption{Boxplots of segmentation quality metrics across brain structures for SynthSeg (left side) and SamSeg (right side). Metrics include Dice coefficient, Intersection over Union, 95th percentile Hausdorff distance, and Normalized Mutual Information.}
    \label{fig:segmentation_metrics}
\end{figure}

Segmentation performance varied substantially between SynthSeg and SamSeg across brain structures (Figure \ref{fig:segmentation_metrics}). Overall, SynthSeg achieved higher agreement with the reference segmentations across all four quality metrics. For the Dice coefficient and IoU, SynthSeg consistently reached values above 0.8 for most structures, while SamSeg displayed greater variability and lower medians, particularly for small subcortical nuclei (e.g., accumbens area, amygdala). Regarding NMI, SynthSeg  demonstrated higher and more stable scores across structures, whereas SamSeg results were more dispersed. The HD95 distances showed low values for SynthSeg across nearly all structures, reflecting close spatial alignment to the reference. In contrast, SamSeg exhibited markedly higher HD95 values with broader distributions, especially for ventricular and deep grey matter regions. Although SynthSeg outperformed SamSeg overall, smaller subcortical regions such as the accumbens area and the inferior lateral ventricle remained comparatively challenging, showing lower Dice and greater variability than larger cortical or ventricular structures.

\subsubsection{Main takeaway}
SynthSeg consistently demonstrated superior segmentation accuracy compared to SamSeg in all evaluated brain structures and quality metrics.

\subsection{Evaluation of Left/Right Differences Relative to Segmentation Error}

To evaluate whether the left and right hemispheric structures can be pooled for subsequent analyses, we compared differences within the subject in volume and FD. We compared left-right differences for the expert annotated data and for most structures, left–right differences in volume were ranged between -2.68 mL to +1.79 mL, or in terms of percentage from -7.5\% to +17.5\%. Large effect sizes in differences were however, only found for Cerebellum-White-Matter, Putamen and Thalamus. For FD, differences between the left and right side where between $\pm$0.02 ($\pm$1\%), except for the Hippocampus (-0.04 (-2.3\%)) and the Lateral Ventricle (-0.21 (-12.3\%)). Furthermore, no large effect sizes were found in left-right differences.

The left-right differences were compared to the differences between the annotated BOB data and results from SynthSeg, the best performing segmentation method (Figure \ref{fig:segmentation_metrics}). Segmentation-related differences were markedly larger, with absolute biases ranging from 0.07 to 27.61 mL in volume (6.1\% to +72.2\% relative to mean structure size). All left–right volume discrepancies were smaller than segmentation errors. For FD, segmentation-related differences ranged from 0.00 to 0.16 (0.0\% to 11.8\%). For three (Hippocampus, Lateral Ventricle and Pallidum) out of 13 structures, side-differences in FD were greater compared to the differences resulting from different segmentations. 

\subsubsection{Main takeaway}

These findings indicate that, although regional asymmetries are detectable in specific structures, they are consistently smaller than segmentation-related variability. Thus, the left and right sides can be justifiably pooled for group-level analyses without introducing systematic bias beyond the uncertainty inherent in segmentation.

\subsection{Volume differences}

Across most brain regions, statistically significant differences in volumetric estimates were observed between expert BOB segmentations, SynthSeg (mean +4\% [range -28\% - 71\%), and SamSeg (mean +76\% [range -12\% - 190\%]) (see Table \ref{tab:volume}). However, the magnitude of differences varied substantially and not all statistically significant results corresponded to large effect sizes. 
For SynthSeg, volumetric estimates were generally close to expert annotations, with differences ranging from +6\% to +16\% in larger structures (e.g., cerebral cortex, cerebellum) and -10\% to -20\% in subcortical nuclei (e.g., hippocampus, thalamus). Most effect sizes were small-to-moderate ($d < 0.8$), indicating that deviations were limited in practical terms. 
In contrast, SamSeg showed pronounced overestimation, particularly in ventricular volumes (e.g., lateral ventricles: +130\%, 4th ventricle: +170\%) and whole-brain tissue (approximately +190\%). These differences corresponded to very large effect sizes (e.g., Cohen’s d = -8.045 for whole-brain volume), highlighting systematic deviations from the expert reference.

\begin{table}[ht]
\centering
\caption{Regional brain volumes (mean $\pm$ SD, in mL) obtained from expert annotations in the BOB dataset compared with SynthSeg and SamSeg outputs. Shown are group means, sample size (n), ANOVA p-values, as well as pairwise comparisons (p-values and Cohen’s d effect sizes).}
\label{tab:volume}
\resizebox{\textwidth}{!}{%
\begin{tabular}{lccccccccccc}
\toprule
& BOB & SynthSeg & SamSeg & n & ANOVA &
\multicolumn{2}{c}{BOB–SynthSeg} &
\multicolumn{2}{c}{BOB–SamSeg} &
\multicolumn{2}{c}{SynthSeg–SamSeg} \\
\cmidrule(lr){7-8} \cmidrule(lr){9-10} \cmidrule(lr){11-12}
Region &  &  &  &  &  
p-val &
p-value & effect size &
p-value & effect size &
p-value & effect size \\
\midrule
3rd-Ventricle & 0.64 $\pm$ 0.14 & 0.54 $\pm$ 0.18 & 1.49 $\pm$ 0.65 & 70 & 0.000 & 0.000 & 0.600 & 0.000 & -1.793 & 0.000 & -1.973 \\
4th-Ventricle & 0.76 $\pm$ 0.17 & 0.57 $\pm$ 0.21 & 2.08 $\pm$ 0.91 & 70 & 0.000 & 0.000 & 1.021 & 0.000 & -2.028 & 0.000 & -2.304 \\
Accumbens-area & 0.32 $\pm$ 0.10 & 0.54 $\pm$ 0.12 & 0.43 $\pm$ 0.10 & 142 & 0.000 & 0.000 & -2.157 & 0.000 & -1.214 & 0.000 & 1.117 \\
Amygdala & 0.79 $\pm$ 0.18 & 1.05 $\pm$ 0.21 & 1.58 $\pm$ 0.32 & 142 & 0.000 & 0.000 & -1.376 & 0.000 & -3.165 & 0.000 & -2.022 \\
Brain-Stem & 9.81 $\pm$ 1.38 & 11.06 $\pm$ 2.45 & 15.41 $\pm$ 3.38 & 71 & 0.000 & 0.000 & -0.627 & 0.000 & -2.166 & 0.000 & -1.473 \\
Caudate & 2.68 $\pm$ 0.61 & 2.35 $\pm$ 0.59 & 2.37 $\pm$ 0.39 & 141 & 0.000 & 0.000 & 0.567 & 0.000 & 0.620 & 1.000 & -0.070 \\
Cerebellum-Cortex & 28.26 $\pm$ 7.46 & 33.52 $\pm$ 8.96 & 46.04 $\pm$ 10.22 & 142 & 0.000 & 0.000 & -0.643 & 0.000 & -2.005 & 0.000 & -1.323 \\
Cerebellum-White-Matter & 5.96 $\pm$ 2.37 & 4.43 $\pm$ 1.41 & 12.35 $\pm$ 2.29 & 142 & 0.000 & 0.000 & 0.791 & 0.000 & -2.783 & 0.000 & -4.237 \\
Cerebral-Cortex & 157.10 $\pm$ 37.96 & 183.81 $\pm$ 27.95 & 192.86 $\pm$ 21.86 & 142 & 0.000 & 0.000 & -0.801 & 0.000 & -1.157 & 0.000 & -0.362 \\
Cerebral-White-Matter & 129.06 $\pm$ 21.48 & 120.00 $\pm$ 18.96 & 243.12 $\pm$ 30.66 & 142 & 0.000 & 0.000 & 0.450 & 0.000 & -4.328 & 0.000 & -4.835 \\
Hippocampus & 1.77 $\pm$ 0.41 & 2.37 $\pm$ 0.52 & 3.48 $\pm$ 0.45 & 142 & 0.000 & 0.000 & -1.317 & 0.000 & -4.072 & 0.000 & -2.365 \\
Inf-Lat-Vent & 0.35 $\pm$ 0.16 & 0.25 $\pm$ 0.09 & 0.52 $\pm$ 0.22 & 142 & 0.000 & 0.000 & 0.901 & 0.000 & -0.972 & 0.000 & -1.697 \\
Lateral-Ventricle & 4.47 $\pm$ 1.58 & 3.48 $\pm$ 1.60 & 10.28 $\pm$ 1.70 & 140 & 0.000 & 0.000 & 0.681 & 0.000 & -4.050 & 0.000 & -4.579 \\
Pallidum & 1.14 $\pm$ 0.36 & 1.07 $\pm$ 0.28 & 1.24 $\pm$ 0.14 & 142 & 0.000 & 0.004 & 0.217 & 0.073 & -0.360 & 0.000 & -0.751 \\
Putamen & 3.24 $\pm$ 0.59 & 3.74 $\pm$ 0.64 & 5.55 $\pm$ 0.85 & 142 & 0.000 & 0.000 & -0.835 & 0.000 & -3.235 & 0.000 & -2.442 \\
Thalamus & 6.50 $\pm$ 0.65 & 5.70 $\pm$ 0.50 & 7.50 $\pm$ 1.51 & 140 & 0.000 & 0.000 & 1.459 & 0.000 & -0.874 & 0.000 & -1.621 \\
VentralDC & 2.01 $\pm$ 0.82 & 2.53 $\pm$ 0.40 & 2.94 $\pm$ 0.65 & 71 & 0.000 & 0.000 & -0.796 & 0.000 & -1.260 & 0.000 & -0.774 \\
Whole-Brain (excl. CSF) & 700.70 $\pm$ 124.98 & 741.63 $\pm$ 120.12 & 2028.57 $\pm$ 197.13 & 71 & 0.000 & 0.000 & -0.334 & 0.000 & -8.045 & 0.000 & -7.884 \\
\bottomrule
\end{tabular}%
}
\end{table}

\subsubsection{Main takeaway}

Overall, SynthSeg provided volumetric estimates largely aligned with the manual reference, with only modest systematic biases. Given its superior performance (Figure \ref{fig:segmentation_metrics}, Table \ref{tab:volume}), the further analysis focused on the comparison between the expert annotations (from BOB dataset) and SynthSeg.

\subsection{Segmentation quality over age}

\begin{figure}
    \centering
    \includegraphics[width=1\linewidth]{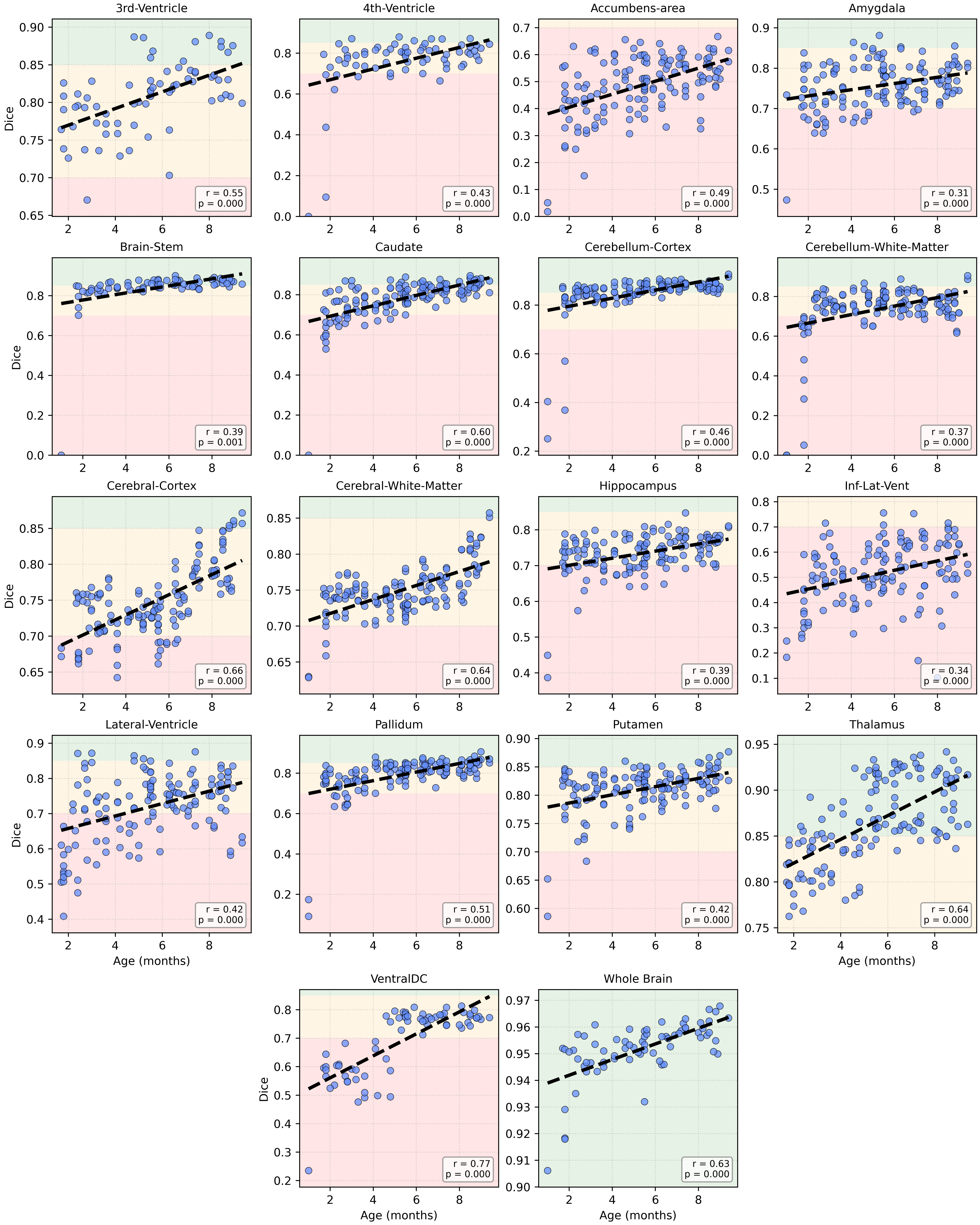}
    \caption{Age-related effects on SynthSeg segmentation accuracy. Scatterplots show Dice coefficients across age (in months) for individual brain structures with fitted regression lines (dashed). Background shading indicates quality thresholds (red: poor, orange: moderate, green: excellent). Pearson correlation coefficients ($r$) and p-values are shown per panel.}
    \label{fig:dice_age}
\end{figure}

In the next step, SynthSeg's segmentation quality was assessed in relation to age (Figure \ref{fig:dice_age}). Across most regions, Dice coefficients increased with age, reflecting improved segmentation accuracy in older infants. Correlations with age were moderate to strong (range 0.31-0.77), and were particularly pronounced in the thalamus ($r = 0.64$), cerebral cortex ($r = 0.66$), cerebral white matter ($r = 0.64$), and ventral diencephalon ($r = 0.77$). These structures showed clear age-related improvements, with many cases exceeding the excellent Dice threshold by 6–9 months of age. In contrast, smaller subcortical regions such as the amygdala, hippocampus, and inferior lateral ventricle showed weaker correlations ($r = 0.31-0.39$) and greater inter-individual variability, indicating persistent segmentation challenges despite age-related improvements. Similar results were also observed for IoU and NMI (positive correlations with age), and HD95 (negative correlations with age) in all regions.

\subsubsection{Main takeaway}
Segmentation accuracy increased with age, indicating more reliable SynthSeg performance in older infants.

\subsection{Fractal dimension}

FD values derived from expert annotations and SynthSeg are summarised in Table ~\ref{tab:fd}. Significant differences in FD were observed for most structures, with the largest effects seen in the thalamus ($d$ = 1.47, $p$ < 0.001), cerebellar cortex ($d$ = –0.97, $p$ < 0.001), and cerebral cortex ($d$ = –0.93, $p$ < 0.001). In contrast, no significant differences were found for the brain stem, lateral ventricles, or pallidum. For the whole brain (excluding CSF), the difference in FD was small with only a moderate effect size ($d$ = –0.50, $p$ < 0.001).

\begin{table}[ht]
\centering
\small
\caption{Fractal dimension (FD) of whole-brain and segmented structures obtained from expert annotations (BOB) and SynthSeg. Values are shown as mean $\pm$ SD. Paired $t$-tests were used to compare FD between methods, and effect sizes are reported as Cohen’s $d$.}
\label{tab:fd}
\begin{tabular}{lllrlr}
\toprule
name & FD (BOB) & FD (SynthSeg) & n & p-val & d \\
\midrule
3rd-Ventricle & 1.74 $\pm$ 0.11 & 1.63 $\pm$ 0.23 & 70 & 0.000 & 0.64 \\
4th-Ventricle & 1.67 $\pm$ 0.15 & 1.50 $\pm$ 0.40 & 70 & 0.001 & 0.41 \\
Accumbens-area & 1.61 $\pm$ 0.08 & 1.71 $\pm$ 0.17 & 142 & 0.000 & -0.46 \\
Amygdala & 1.84 $\pm$ 0.12 & 1.90 $\pm$ 0.15 & 142 & 0.000 & -0.41 \\
Brain-Stem & 2.29 $\pm$ 0.02 & 2.27 $\pm$ 0.22 & 71 & 0.562 & 0.07 \\
Caudate & 2.01 $\pm$ 0.04 & 1.89 $\pm$ 0.30 & 141 & 0.000 & 0.42 \\
Cerebellum-Cortex & 2.44 $\pm$ 0.04 & 2.47 $\pm$ 0.05 & 142 & 0.000 & -0.97 \\
Cerebellum-White-Matter & 2.08 $\pm$ 0.06 & 2.04 $\pm$ 0.17 & 142 & 0.002 & 0.26 \\
Cerebral-Cortex & 2.47 $\pm$ 0.05 & 2.51 $\pm$ 0.02 & 142 & 0.000 & -0.93 \\
Cerebral-White-Matter & 2.44 $\pm$ 0.04 & 2.45 $\pm$ 0.01 & 142 & 0.000 & -0.45 \\
Hippocampus & 1.93 $\pm$ 0.08 & 1.96 $\pm$ 0.14 & 142 & 0.005 & -0.24 \\
Inf-Lat-Vent & 1.36 $\pm$ 0.18 & 1.20 $\pm$ 0.15 & 142 & 0.000 & 0.81 \\
Lateral-Ventricle & 1.72 $\pm$ 0.23 & 1.72 $\pm$ 0.26 & 140 & 0.893 & 0.01 \\
Pallidum & 1.89 $\pm$ 0.17 & 1.89 $\pm$ 0.16 & 142 & 0.913 & 0.01 \\
Putamen & 2.08 $\pm$ 0.12 & 2.11 $\pm$ 0.10 & 142 & 0.009 & -0.22 \\
Thalamus & 2.28 $\pm$ 0.02 & 2.26 $\pm$ 0.02 & 140 & 0.000 & 1.47 \\
VentralDC & 1.85 $\pm$ 0.19 & 1.95 $\pm$ 0.15 & 71 & 0.001 & -0.42 \\
Whole brain (excl. CSF) & 2.77 $\pm$ 0.02 & 2.77 $\pm$ 0.01 & 71 & 0.000 & -0.50 \\
\bottomrule
\end{tabular}
\end{table}

\begin{figure}
    \centering
    \includegraphics[width=1\linewidth]{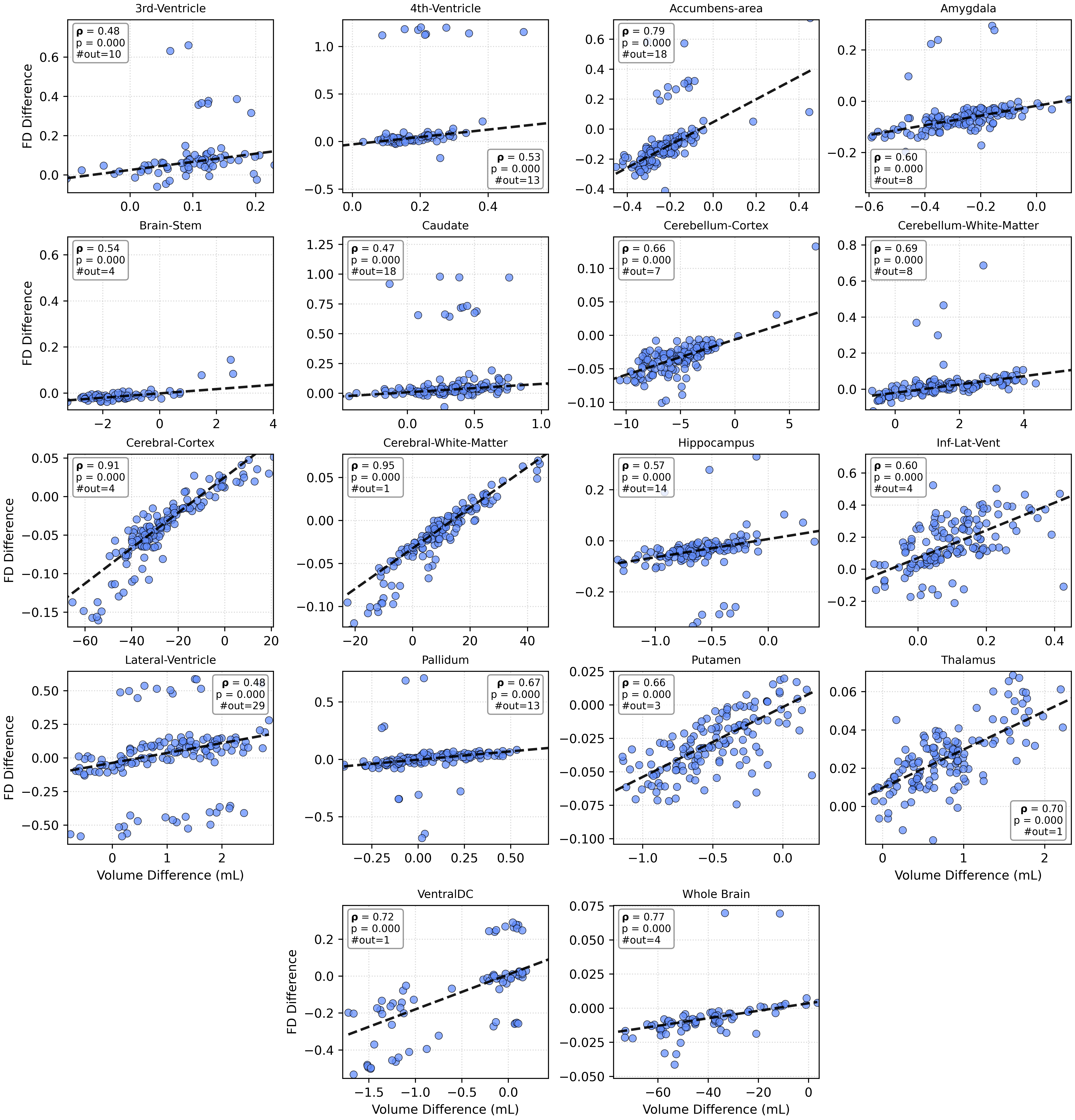}
    \caption{Fractal dimension (FD) differences over volume differences for individual brain structures. Each panel shows a robust regression line (Theil–Sen estimator, dashed) and the Spearman correlation coefficient ($\rho$) 
    and the associated $p$-value. Outlier counts (\#out) are indicated based on a median absolute deviation (MAD) threshold (|$z$| > 3.5), but all data points remain included in both regression and correlation analyses. For clearer visualisation, axes are limited to the 1st–99th percentile range of the data within each structure.}
    \label{fig:fd_vol}
\end{figure}

Across all brain regions, volume differences were positively correlated with FD differences (Figure ~\ref{fig:fd_vol}). 

\subsubsection{Main takeaway}
Most brain structures showed significant differences in FD between SynthSeg and expert annotations, and FD deviations were positively correlated with corresponding volume differences.

\subsection{Limits of agreement}

For volumetric measures, biases ranged from –9.07 $\pm$ 26.40 mL (–6.5 $\pm$ 18.6\%) in cerebral white matter to 40.93 $\pm$ 36.01 mL (6.2 $\pm$ 6.2\%) for the whole brain. The largest relative bias was observed in the accumbens area (83.9 $\pm$ 107.8\%).
For FD, biases ranged from –0.16 $\pm$ 0.38 (–10.2 $\pm$ 35.2\%) in the inferior lateral ventricle to 0.10 $\pm$ 0.45 (6.3 $\pm$ 26.5\%) in the ventral diencephalon.

\begin{table}[ht]
\centering
\caption{Bland–Altman analysis of SynthSeg- and BOB-derived brain volumes and fractal dimensions (FD) across individual structures. For each region, the bias (mean difference, SynthSeg – BOB) and the corresponding 95\% limits of agreement (LoA) are reported in both absolute units (mL or raw FD) and relative values (\%).}
\label{tab:bland_altman}
\begin{adjustbox}{max width=\textwidth}
\begin{tabular}{lcccc}
\toprule
\textbf{Structure} &
\makecell[c]{\textbf{Volume}\\(Bias $\pm$ LoA)\\{[mL]}} &
\makecell[c]{\textbf{Volume}\\(Bias $\pm$ LoA)\\{[\%]}} &
\makecell[c]{\textbf{FD}\\(Bias $\pm$ LoA)} &
\makecell[c]{\textbf{FD}\\(Bias $\pm$ LoA)\\{[\%]}} \\
\midrule
3rd-Ventricle & -0.10 $\pm$ 0.13 & -16.6 $\pm$ 21.0 & -0.11 $\pm$ 0.33 & -6.4 $\pm$ 20.2 \\
4th-Ventricle & -0.19 $\pm$ 0.20 & -27.0 $\pm$ 33.1 & -0.17 $\pm$ 0.82 & -8.9 $\pm$ 61.4 \\
Accumbens-area & 0.23 $\pm$ 0.27 & 83.9 $\pm$ 107.8 & 0.09 $\pm$ 0.40 & 6.1 $\pm$ 24.3 \\
Amygdala & 0.26 $\pm$ 0.27 & 36.1 $\pm$ 43.1 & 0.06 $\pm$ 0.29 & 3.8 $\pm$ 23.2 \\
Brain-Stem & 1.25 $\pm$ 2.84 & 11.6 $\pm$ 35.9 & -0.02 $\pm$ 0.43 & -0.7 $\pm$ 19.0 \\
Caudate & -0.34 $\pm$ 0.55 & -12.7 $\pm$ 26.1 & -0.12 $\pm$ 0.54 & -6.0 $\pm$ 28.0 \\
Cerebellum-Cortex & 5.26 $\pm$ 5.87 & 18.4 $\pm$ 28.6 & 0.03 $\pm$ 0.07 & 1.4 $\pm$ 2.9 \\
Cerebellum-White-Matter & -1.53 $\pm$ 2.59 & -22.0 $\pm$ 42.2 & -0.04 $\pm$ 0.30 & -2.0 $\pm$ 15.4 \\
Cerebral-Cortex & 26.71 $\pm$ 33.49 & 19.7 $\pm$ 26.7 & 0.04 $\pm$ 0.09 & 1.7 $\pm$ 3.7 \\
Cerebral-White-Matter & -9.07 $\pm$ 26.40 & -6.5 $\pm$ 18.6 & 0.02 $\pm$ 0.08 & 0.7 $\pm$ 3.2 \\
Hippocampus & 0.60 $\pm$ 0.61 & 34.9 $\pm$ 41.8 & 0.03 $\pm$ 0.27 & 1.8 $\pm$ 15.1 \\
Inf-Lat-Vent & -0.10 $\pm$ 0.24 & -19.6 $\pm$ 89.1 & -0.16 $\pm$ 0.38 & -10.2 $\pm$ 35.2 \\
Lateral-Ventricle & -0.99 $\pm$ 1.78 & -22.3 $\pm$ 39.2 & -0.00 $\pm$ 0.52 & 0.8 $\pm$ 33.2 \\
Pallidum & -0.07 $\pm$ 0.43 & -2.0 $\pm$ 41.8 & -0.00 $\pm$ 0.39 & 1.3 $\pm$ 38.8 \\
Putamen & 0.50 $\pm$ 0.66 & 16.2 $\pm$ 25.5 & 0.03 $\pm$ 0.28 & 2.2 $\pm$ 29.0 \\
Thalamus & -0.80 $\pm$ 1.11 & -11.9 $\pm$ 15.2 & -0.03 $\pm$ 0.03 & -1.1 $\pm$ 1.5 \\
VentralDC & 0.51 $\pm$ 1.31 & 48.4 $\pm$ 130.3 & 0.10 $\pm$ 0.45 & 6.3 $\pm$ 26.5 \\
Whole Brain (excl. CSF) & 40.93 $\pm$ 36.01 & 6.2 $\pm$ 6.2 & 0.01 $\pm$ 0.03 & 0.3 $\pm$ 1.2 \\
\bottomrule
\end{tabular}
\end{adjustbox}
\end{table}

\subsubsection{Main takeaway}
These bias and LoA estimates serve as benchmarks for segmentation uncertainty, allowing future investigations to assess whether observed group or developmental differences surpass the variability inherent to automated segmentation.

\section{Discussion}


A summary of the main takeaway from the results can be found in Table \ref{tab:main_takeaways}.
\begin{table}[h!]
\centering
\footnotesize
\caption{Overview of main analyses and corresponding key findings.}
\label{tab:main_takeaways}
\resizebox{0.8\textwidth}{!}{%
\begin{tabularx}{\textwidth}{@{}p{0.37\textwidth}X@{}}
\toprule
\textbf{Experiment / analysis} & \textbf{Main takeaway} \\
\midrule
Comparison of segmentation labels quality between SynthSeg and SamSeg &
Across all brain structures and standard quality metrics (Dice coefficient, Intersection over Union, 95th percentile Hausdorff distance, and Normalised Mutual Information), SynthSeg demonstrated consistently superior segmentation accuracy and stability compared with SamSeg, which showed lower and more variable performance, particularly in small subcortical and ventricular regions. \\[0.5em]

Evaluation of hemispheric symmetry relative to segmentation error &
Within-subject left--right differences in volume and fractal dimension were present but consistently smaller than the discrepancies introduced by automated segmentation. Consequently, pooling left and right hemispheres for group-level analyses is justified, as it does not add systematic bias beyond the intrinsic segmentation-related variability. \\[0.5em]

Regional and whole-brain volume differences between methods &
SynthSeg produced volumetric estimates that were closely aligned with expert annotations, with only modest systematic biases across structures. In contrast, SamSeg systematically and substantially overestimated ventricular and whole-brain volumes, leading to very large effect sizes; therefore, subsequent volumetric and downstream analyses were based on SynthSeg for its quantitative volumetric outputs. \\[0.5em]

Segmentation quality as a function of age &
Segmentation accuracy for SynthSeg improved with increasing infant age across most regions, reflected by higher Dice, IoU and NMI and lower HD95 in older infants. By approximately 6--9~months, many structures reached excellent agreement with the reference, although small regions (i.e. Accumbens area and inferior lateral ventricle) remained comparatively challenging, indicating more reliable performance in older infants. \\[0.5em]

Impact of segmentation on fractal dimension estimates &
Most brain structures exhibited significant differences in fractal dimension between SynthSeg and expert segmentations. Moreover, deviations in fractal dimension were positively correlated with corresponding volume differences, indicating that segmentation bias propagates directly into higher-order morphological measures of structural complexity. \\[0.5em]

Limits of agreement for volume and fractal dimension &
Bland--Altman analyses established quantitative biases and 95\% limits of agreement for both volume and fractal dimension across structures for infants aged 1 to 9 months. These benchmarks delineate the magnitude of segmentation-related uncertainty and provide a reference against which future group or developmental differences can be judged to determine whether they exceed the variability inherent to automated segmentation. \\
\bottomrule
\end{tabularx}}
\end{table}

In this study, we systematically evaluated how the choice of automated segmentation pipeline influences estimates of brain volume and fractal dimension (FD), a measure of structural complexity, in neonatal and early-infant MRI. Focussing on SynthSeg and SamSeg as two widely used but methodologically distinct frameworks, and benchmarking them against expert annotations, we show that segmentation error is not a negligible preprocessing detail but a primary determinant of downstream morphometric estimates. Importantly, we demonstrate that biases at the segmentation stage propagate directly into FD, meaning that even subtle systematic errors in tissue delineation can materially affect higher-order descriptors of cortical and subcortical organisation.

Our analyses establish SynthSeg as the most reliable of the tools evaluated for infant MRI, extending previous work in adults~\cite{billot2023synthseg} to a much younger and technically more challenging population. SynthSeg achieved consistently higher overlap and distance-based metrics than SamSeg across structures and ages, and produced volumetric estimates that aligned more closely with expert labels. By contrast, SamSeg showed larger and more variable discrepancies, including systematic overestimation of key compartments, and three additional adult-optimised pipelines (FSL FAST~\cite{zhang2002segmentation}, FastSurfer~\cite{henschel2020fastsurfer}, and the MONAI UNesT model~\cite{cardoso2022monai}) did not produce anatomically plausible segmentations in this age range. Together, these findings highlight that widely used adult pipelines cannot be assumed to generalise to early brain development and that model priors and training data must be carefully matched to the developmental stage of interest. Segmentation quality also improved with age, reflecting early brain maturation. Myelination begins before birth and progresses in a central-to-peripheral, caudal-to-rostral pattern~\cite{carmody2004quantitative,parazzini2002terminal,welker2012assessment}, gradually enhancing tissue contrast and anatomical definition.

To contextualise the observed biases and limits of agreement (LoA) for volumetric and FD accuracy, we compared them with effect sizes reported in the literature. In infants with intrauterine growth restriction, preterm birth, and term-born controls at 12 $\pm$ 2 months of corrected age, Esteban et al.\cite{esteban2010fractal} reported no significant volumetric differences in grey matter, white matter, or total brain volume, with mean group differences remaining below 40 mL. Our volumetric LoA, $\pm$36 mL for whole brain, $\pm$33 mL for cerebral cortex, and $\pm$26 mL for cerebral white matter, were therefore of a similar magnitude, indicating that segmentation uncertainty can match the scale of subtle between-group effects in early infancy. In older paediatric cohorts with Attention Deficit Hyperactivity Disorder (ADHD), volumetric reductions of –4.1\% in total cerebrum, –3.7\% in the grey matter, –4.8\% in the white matter, and –6.7\% in the caudate have been reported.\cite{castellanos2002developmental} These values are comparable to our LoA for the whole brain (6.2\% $\pm$ 6.2\%) but are far smaller than cortical LoA (19.7\% $\pm$ 26.7\%)  and white matter (–6.5\% $\pm$ 18.6\%). Small subcortical and intracranial differences of 1–5\% in the youngest ENIGMA-ADHD subgroup\cite{hoogman2017subcortical} lie within, or are exceeded by, our segmentation-related variability. By contrast, early-autism cohorts show large volumetric increases, 10\% in total brain volume, 12\% in cortical grey matter, 18\% in white matter, and up to 39\% in cerebellar white matter, all similar to or exceeding our LoA, except for cortical grey matter\cite{courchesne2001unusual}. For FD, we only considered studies based on three-dimensional FD calculations. Esteban et al.\cite{esteban2010fractal} reported differences between-groups of 0.007 (grey matter), 0.015 (skeletonised grey matter), and 0.046 (white matter). In comparison, our LoA were $\pm$0.09 for cerebral cortex and $\pm$0.08 for white matter, exceeding or matching these effects. In adults, Farahibozorg et al.\cite{Farahibozorg2015} reported age- and sex-related variations of ~0.01–0.03 in white-matter FD, and Di Marco et al.\cite{DiMarco2024} observed FD reductions of 0.049 in cortical grey matter in Alzheimer’s disease, which are both smaller than our segmentation-related variability. Much larger FD alterations have been observed in spinocerebellar ataxia type 2, with group differences of ~0.08 in cerebellar cortex, ~0.48 in cerebellar white matter, and ~0.04 in cerebral cortex.\cite{marzi2018structural} For these regions, our segmentation LoA were smaller or comparable for cerebellar measures but exceeded disease effects for cerebral cortex. Taken together, these comparisons indicate that current segmentation accuracy may be insufficient to detect subtle group differences, as segmentation uncertainty can match or exceed the magnitude of effects reported in mildly affected cohorts. Only in the presence of pronounced volumetric or FD alterations is segmentation uncertainty unlikely to obscure true group differences.

Several limitations should be considered when interpreting the present findings. First, the analyses were based on the Baby Open Brains (BOB) dataset, which includes a limited age range and sample size, potentially restricting generalisability across the broader developmental spectrum of infancy and early childhood. However, because segmentation accuracy improved with increasing age, the main limitation is confined to the youngest newborns, where reduced tissue contrast remains the major challenge. Second, both SynthSeg and SamSeg were applied using their publicly available models without retraining or fine-tuning for infant MRI. This choice was deliberate in comparing their out-of-the-box performance and in assessing robustness under real-world conditions where retraining data are scarce and standardised neonatal ground truths are limited. However, domain-adapted models may achieve higher accuracy once suitable training data become available. Third, FD was computed on three-dimensional binary masks to ensure methodological consistency across structures and segmentation methods. Although this approach facilitates comparability, it is inherently sensitive to voxel resolution, partial-volume effects, and boundary noise, particularly in small subcortical regions. Future work should include larger and longitudinal datasets, evaluate retrained or hybrid segmentation models, and explore harmonisation strategies to improve the reproducibility of both volumetric and fractal-based measures in developmental studies.

\section{Conclusion}

This study demonstrates that segmentation accuracy strongly influences estimates of brain volume and fractal dimension in paediatric MRI. SynthSeg consistently outperformed SamSeg and achieved robust results even in the youngest infants, although segmentation of the small regions (i.e. Accumbens area and inferior lateral ventricle) remains challenging. The observed age-related improvement in segmentation quality aligns with ongoing myelination and increasing tissue contrast during early development. Comparison with previous literature indicates that segmentation-related limits of agreement for volume overlap with the magnitude of many reported group differences. As a result, the present pipeline is likely adequate for identifying robust, large-scale effects, but may lack the precision required to reliably quantify smaller or more nuanced volumetric alterations. For fractal dimension, the limits of agreement tend to exceed most group differences reported in developmental and clinical cohorts, suggesting that small fractal dimension effects may fall within the range of segmentation uncertainty. Importantly, the present study focuses on very young infants (1–9 months), an age range in which low tissue contrast and rapid neurodevelopment pose particular challenges for automated segmentation. Given that segmentation accuracy generally improves with increasing age and cortical maturation, the degree to which these limitations persist beyond infancy remains an open question. Thus, while the current pipeline may be constrained in early infancy, its suitability for older developmental stages may be substantially greater. Overall, these findings highlight the importance of accounting for segmentation accuracy when interpreting quantitative MRI biomarkers and support the use of SynthSeg-based volumetric measures as reliable indicators of early brain development.


\bibliographystyle{elsarticle-num} 
\bibliography{references}
\end{document}